\documentclass[11pt,a4paper]{article}

\usepackage[utf8]{inputenc}
\usepackage[T1]{fontenc}
\usepackage{lmodern}
\usepackage[margin=1in]{geometry}
\usepackage{amsmath,amssymb,amsfonts}
\usepackage{graphicx}
\usepackage{booktabs}
\usepackage{multirow}
\usepackage{colortbl}
\usepackage{xcolor}
\usepackage{hyperref}
\usepackage{url}
\usepackage{natbib}
\usepackage{pifont}
\usepackage{placeins}
\usepackage{float}
\usepackage{tikz}
\usetikzlibrary{shapes,arrows.meta,positioning,fit,backgrounds,calc}

\hypersetup{
    colorlinks=true,
    linkcolor=blue,
    citecolor=blue,
    urlcolor=blue
}

\title{PROTEUS: SLA-Aware Routing via Lagrangian RL\\for Multi-LLM Serving Systems}

\author{
Amit Singh Bhatti\thanks{Phi Labs, Quantiphi, Boston, USA. Email: amit.bhatti@quantiphi.com} \and
Vishal Vaddina\thanks{Phi Labs, Quantiphi, Boston, USA. Email: vishal.vaddina@quantiphi.com} \and
Dagnachew Birru\thanks{Phi Labs, Quantiphi, Boston, USA. Email: dagnachew.birru@quantiphi.com}
}

\date{}

\begin{document}

\maketitle

\begin{abstract}
Production LLM deployments serve diverse workloads where cost and quality requirements vary by customer tier, time of day, and query criticality. Model serving systems accept latency service-level objectives (SLOs) directly. LLM routers do not. They force operators to tune parameters offline and guess what accuracy might result. The relationship between parameters and outcomes is indirect, non-monotonic, and dataset-dependent. Operators need to specify accuracy targets, not infer them from opaque settings. We present \textbf{PROTEUS} (\textbf{P}olymorphic \textbf{R}outer for \textbf{O}perational \textbf{T}arget \textbf{E}nforce\-ment with \textbf{U}nified \textbf{S}LA), a router that accepts accuracy targets $\tau$ as runtime input. PROTEUS uses Lagrangian dual control. A learned dual variable $\lambda$ tracks constraint violations during training and conditions the policy network. This lets the router translate specified $\tau$ values into routing decisions that satisfy them. A single trained model serves the full accuracy spectrum without retraining. We evaluate on RouterBench (11 models, 405K queries) and SPROUT (14 models, 45K queries). PROTEUS achieves consistent floor compliance where accuracy meets or exceeds $\tau$. The target-response correlation reaches 0.97 to 0.98. The closest baseline, OmniRouter, meets floors only 22\% of the time despite also using Lagrangian optimization. PROTEUS operates across $\tau \in [0.85, 0.95]$ from a single model. On RouterBench it achieves 90.1\% accuracy, within 1.3\% of oracle. On SPROUT it achieves 94.0\% accuracy, within 4.6\% of oracle. Cost savings reach 89.8\% versus the best fixed model.
\end{abstract}

\section{Introduction}
\label{sec:intro}

Production LLM serving systems must balance conflicting objectives, maximizing response quality while minimizing operational costs under varying workload requirements. Existing routing approaches optimize this tradeoff indirectly through tunable parameters such as confidence thresholds, routing percentages, or constraint weights, but these parameters have a non-monotonic, dataset-dependent relationship with actual quality outcomes~\citep{routerbench2024,routereval2025}. Operators cannot directly specify quality requirements as service-level objectives (SLOs), even though traditional serving systems let you specify latency or throughput targets directly~\citep{clipper2017,infaas2021}. This indirect control becomes operationally infeasible as deployments scale across differentiated service tiers with distinct quality and cost requirements.

Production deployments exhibit heterogeneous quality requirements across workload classes, as evidenced by OpenAI's API offering differentiated service tiers with per-request \texttt{service\_tier} selection~\citep{openai_api}, AWS identifying multi-tenant SaaS architectures as a primary routing use case~\citep{aws_bedrock}, and Microsoft's production traces revealing workload classes with conflicting latency and quality SLAs~\citep{sageserve2025}. In these systems, customer tier metadata or SLA contracts determine quality requirements, where a premium customer subscribing to high-accuracy service expects 95\% quality while an economy customer with cost-optimized access expects 75\% quality. Although these platforms accept latency and availability targets directly, quality requirements remain indirectly controlled, and batch analytics workloads may tolerate lower accuracy for significant cost reduction while customer-facing queries require high reliability regardless of cost. Current routers force operators to infer which parameter settings might achieve desired accuracy levels rather than specifying accuracy targets directly.

LLM routing should solve this problem, given that models span four orders of magnitude in cost~\citep{routerbench2024,helm} and effective routing reduces costs by 80 to 90\%~\citep{carrot2025,hybridllm2024}, but current routers control outcomes indirectly. Cascade methods~\citep{frugalgpt2023,automix2024} route through models until confidence thresholds are met, while learned predictors~\citep{routellm2024,carrot2025,hybridllm2024} require threshold calibration. RouteLLM~\citep{routellm2024} asks operators to ``calibrate thresholds using a sample of queries'' to estimate what accuracy a given parameter produces, meaning operators must deploy separate configurations per tier or retune parameters when requirements change. RL methods~\citep{omnirouter2025,router-r1-2025} optimize with constraint weights, but the relationship between parameters and outcomes remains indirect, non-monotonic, and dataset-dependent~\citep{routerbench2024,routereval2025}.

Model serving systems solved the analogous problem for latency, with Clipper~\citep{clipper2017} accepting latency SLOs directly via additive increase multiplicative decrease (AIMD) control, INFaaS~\citep{infaas2021} accepting accuracy requirements such as 70\% target accuracy specified via the command-line flag \texttt{-A 70} and selecting model variants accordingly, and Proteus~\citep{proteus2024} scaling accuracy reactively under load. However, none of these systems address multi-LLM routing with learned policies, as INFaaS selects among variants of one architecture and Proteus's scaling is system-driven rather than operator-specified. What is missing is a router that accepts accuracy targets directly and learns to satisfy them across heterogeneous LLMs.

This problem is harder than threshold tuning because the policy must generalize across continuous $\tau$ values without memorizing operating points, and static model bucketization fails when query difficulty varies within each tier. For example, a premium customer asking a simple factual question should be routed to a cheap model that satisfies the 95\% target, whereas the same customer asking a complex reasoning question needs an expensive model. The router therefore needs per-query adaptation conditioned on both target accuracy and query difficulty, making a global threshold shift or fixed tier-to-model mapping insufficient.

Our approach uses constrained RL with $\tau$-conditioned networks, where a learned Lagrangian dual variable $\lambda$ enforces constraints during training by penalizing accuracy shortfalls. The policy outputs a continuous quality preference $\mu \in [0,1]$, which allows smooth interpolation across operating points and eliminates manual tuning.

\textbf{Contributions.} We formalize SLA-adaptive routing as constrained optimization with $\tau$ as runtime parameter. We present PROTEUS, which combines $\tau$-conditioned networks with Lagrangian dual control to enable a single trained model to serve the full accuracy spectrum. We demonstrate consistent floor compliance on two benchmarks where baselines fail, while matching or exceeding their standard routing metrics.

\section{Method}
\label{sec:method}

\subsection{Problem Formulation}
A routing system receives queries $x$ and must select from a pool of $K$ models with varying capabilities and costs, where each model $m_i$ has per-query performance $p_i(x) \in [0,1]$ representing the correctness probability and a cost $c_i$ in dollars per query. The operator specifies an accuracy target $\tau$ representing the minimum acceptable accuracy derived from customer SLA contracts or service tier metadata, and the router must minimize cost while meeting this target.
\begin{equation}
\min_{\pi_\theta} \; \mathbb{E}_{x, m \sim \pi_\theta(x,\tau)}[c_m] \quad \text{s.t.} \quad \mathbb{E}_{x, m \sim \pi_\theta(x,\tau)}[p_m(x)] \geq \tau
\label{eq:objective}
\end{equation}
Unlike prior work~\citep{routellm2024,carrot2025,omnirouter2025}, we condition the policy on $\tau$ as runtime input, allowing a single deployed model to serve specified accuracy requirements without retraining or parameter sweeping.

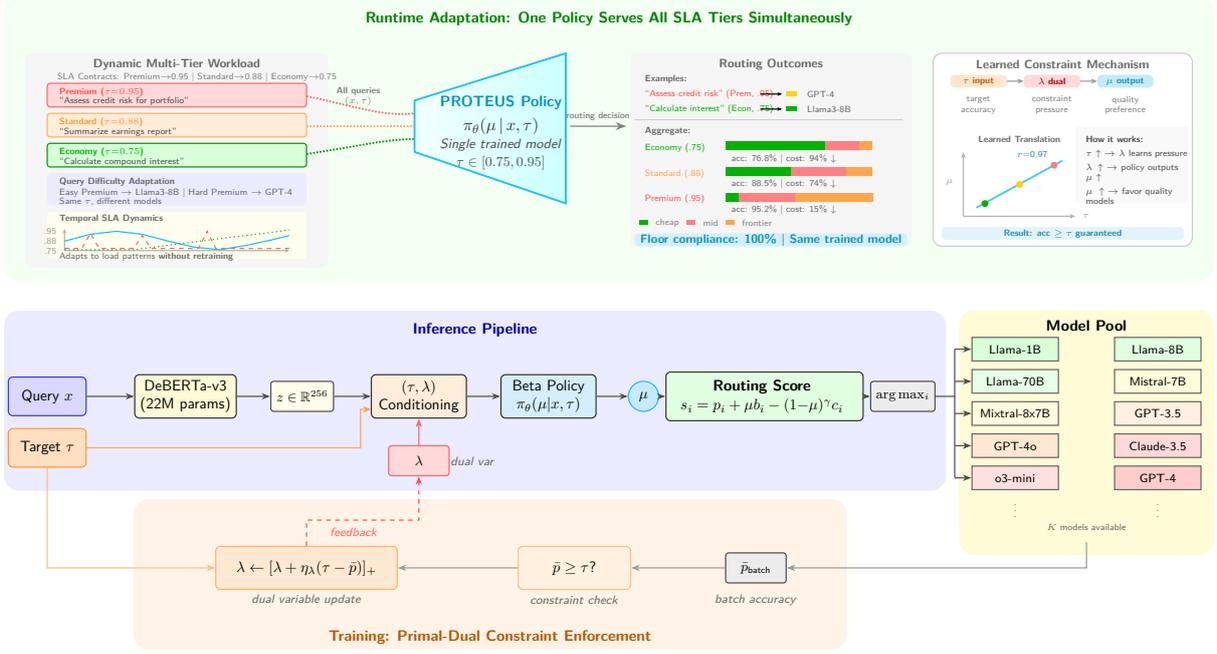
\begin{figure*}[t]
\centering
\resizebox{\textwidth}{!}{%
\begin{tikzpicture}[
    node distance=0.8cm and 1.0cm,
    component/.style={rectangle, draw=black!80, thick, rounded corners=3pt, minimum height=1.0cm, minimum width=2.2cm, align=center, font=\sffamily\small},
    smallcomp/.style={rectangle, draw=black!70, thick, rounded corners=2pt, minimum height=0.7cm, minimum width=1.4cm, align=center, font=\sffamily\footnotesize},
    input/.style={rectangle, draw=blue!80, thick, rounded corners=3pt, minimum height=0.9cm, minimum width=1.8cm, align=center, font=\sffamily\small, fill=blue!15},
    modelbox/.style={rectangle, draw=black!60, thick, minimum height=0.55cm, minimum width=2.4cm, align=center, font=\sffamily\scriptsize, fill=white},
    circlenode/.style={circle, draw=black!80, thick, minimum size=0.7cm, align=center, font=\sffamily\small},
    querybox/.style={rectangle, draw=gray!60, fill=white, minimum width=0.4cm, minimum height=0.35cm, font=\sffamily\tiny, inner sep=1pt},
    arrow/.style={-{Stealth[length=2mm]}, thick, black!70},
    feedbackarrow/.style={-{Stealth[length=2mm]}, thick, dashed, red!70},
    grouplabel/.style={font=\sffamily\small\bfseries},
    complabel/.style={font=\sffamily\scriptsize, black!60},
]

\fill[green!5, rounded corners=10pt] (-1.0,4.9) rectangle (27.0,11.5);
\node[grouplabel, green!50!black] at (13.0,11.0) {Runtime Adaptation: One Policy Serves All SLA Tiers Simultaneously};

\fill[gray!8, rounded corners=4pt] (-0.5,5.2) rectangle (6.5,10.2);
\node[font=\sffamily\scriptsize\bfseries, black!60] at (3.0,9.95) {Dynamic Multi-Tier Workload};

\node[font=\sffamily\tiny, black!50, anchor=west] at (0.1,9.65) {SLA Contracts: Premium$\rightarrow$0.95 | Standard$\rightarrow$0.88 | Economy$\rightarrow$0.75};

\fill[red!15, rounded corners=2pt] (0.0,8.95) rectangle (6.0,9.5);
\draw[red!60, thick, rounded corners=2pt] (0.0,8.95) rectangle (6.0,9.5);
\node[font=\sffamily\tiny\bfseries, red!70, anchor=west] at (0.15,9.3) {Premium ($\tau{=}0.95$)};
\node[font=\sffamily\tiny, black!70, anchor=west, text width=5.6cm] at (0.15,9.08) {``Assess credit risk for portfolio''};

\fill[orange!15, rounded corners=2pt] (0.0,8.25) rectangle (6.0,8.8);
\draw[orange!60, thick, rounded corners=2pt] (0.0,8.25) rectangle (6.0,8.8);
\node[font=\sffamily\tiny\bfseries, orange!80, anchor=west] at (0.15,8.6) {Standard ($\tau{=}0.88$)};
\node[font=\sffamily\tiny, black!70, anchor=west, text width=5.6cm] at (0.15,8.38) {``Summarize earnings report''};

\fill[green!15, rounded corners=2pt] (0.0,7.55) rectangle (6.0,8.1);
\draw[green!60!black, thick, rounded corners=2pt] (0.0,7.55) rectangle (6.0,8.1);
\node[font=\sffamily\tiny\bfseries, green!70!black, anchor=west] at (0.15,7.9) {Economy ($\tau{=}0.75$)};
\node[font=\sffamily\tiny, black!70, anchor=west, text width=5.6cm] at (0.15,7.68) {``Calculate compound interest''};

\fill[blue!8, rounded corners=3pt] (0.0,6.65) rectangle (6.0,7.4);
\node[font=\sffamily\tiny\bfseries, black!60, anchor=west] at (0.15,7.2) {Query Difficulty Adaptation};
\node[font=\sffamily\tiny, black!60, anchor=west] at (0.15,6.95) {Easy Premium $\rightarrow$ Llama3-8B | Hard Premium $\rightarrow$ GPT-4};
\node[font=\sffamily\tiny, black!60, anchor=west, text width=5.6cm] at (0.15,6.75) {Same $\tau$, different models};

\fill[yellow!8, rounded corners=3pt] (0.0,5.4) rectangle (6.0,6.5);
\node[font=\sffamily\tiny\bfseries, black!60, anchor=west] at (0.15,6.35) {Temporal SLA Dynamics};
\draw[-{Stealth[length=1mm]}, thick, black!30] (0.4,5.6) -- (0.4,6.1);
\draw[-{Stealth[length=1mm]}, thick, black!30] (0.4,5.6) -- (5.6,5.6);
\node[font=\sffamily\tiny, black!40, anchor=east] at (0.35,6.05) {.95};
\node[font=\sffamily\tiny, black!40, anchor=east] at (0.35,5.82) {.88};
\node[font=\sffamily\tiny, black!40, anchor=east] at (0.35,5.6) {.75};
\draw[cyan!70, thick] (0.4,5.82) -- (1.0,5.98) -- (1.6,6.05) -- (2.2,5.98) -- (2.8,5.82) -- (3.4,5.68) -- (4.0,5.62) -- (4.6,5.72) -- (5.2,5.85) -- (5.6,5.95);
\draw[red!60, thick, dashed] (0.4,5.65) -- (0.8,5.65) -- (1.0,6.0) -- (1.2,5.65) -- (2.0,5.65) -- (2.2,5.95) -- (2.4,5.65) -- (3.5,5.65) -- (3.7,6.05) -- (3.9,5.65) -- (5.6,5.65);
\draw[green!70!black, thick, dotted] (0.4,5.62) -- (2.0,5.62) -- (3.0,5.75) -- (4.0,5.88) -- (5.0,6.0) -- (5.6,6.08);
\node[font=\sffamily\tiny, black!60, anchor=west, text width=5.4cm] at (0.15,5.45) {Adapts to load patterns \textbf{without retraining}};

\draw[red!60, very thick, densely dotted] (6.0,9.2) .. controls (7.0,9.0) and (7.8,8.7) .. (8.5,8.8);
\draw[orange!70, very thick, densely dotted] (6.0,8.5) .. controls (7.0,8.5) and (7.8,8.5) .. (8.5,8.5);
\draw[green!60!black, very thick, densely dotted] (6.0,7.85) .. controls (7.0,8.0) and (7.8,8.2) .. (8.5,8.2);

\node[font=\sffamily\tiny\bfseries, black!50, align=center] at (7.2,9.2) {All queries\\$(x, \tau)$};


\shade[left color=cyan!5, right color=cyan!20] (8.5,7.9) -- (8.5,9.1) -- (12.0,10.2) -- (12.0,6.7) -- cycle;
\draw[cyan!70, very thick] (8.5,9.1) -- (12.0,10.2) -- (12.0,6.7) -- (8.5,7.9) -- cycle;

\node[font=\sffamily\small\bfseries, cyan!70!black] at (10.5,9.05) {PROTEUS Policy};
\node[font=\sffamily\normalsize, black!80] at (10.5,8.5) {$\pi_\theta(\mu \,|\, x, \tau)$};

\node[font=\sffamily\footnotesize, black!70, align=center] at (10.5,7.85) {\textit{Single trained model}\\[1pt]$\tau \in [0.75, 0.95]$};

\draw[-{Stealth[length=2.5mm]}, thick, black!50] (12.1,8.5) -- (13.4,8.5);
\node[font=\sffamily\tiny, black!60] at (12.75,8.75) {routing decision};

\fill[gray!8, rounded corners=4pt] (13.5,5.7) rectangle (20.0,10.2);
\node[font=\sffamily\scriptsize\bfseries, black!60] at (16.75,9.95) {Routing Outcomes};

\node[font=\sffamily\tiny\bfseries, black!60, anchor=west] at (13.7,9.6) {Examples:};

\node[font=\sffamily\tiny, red!70, anchor=west] at (13.7,9.25) {``Assess credit risk'' (Prem, .95)};
\draw[-{Stealth[length=1mm]}, thick] (16.5,9.25) -- (17.0,9.25);
\fill[yellow!70!orange] (17.1,9.3) rectangle (17.35,9.2);
\node[font=\sffamily\tiny, black!60, anchor=west] at (17.45,9.25) {GPT-4};

\node[font=\sffamily\tiny, green!70!black, anchor=west] at (13.7,8.9) {``Calculate interest'' (Econ, .75)};
\draw[-{Stealth[length=1mm]}, thick] (16.5,8.9) -- (17.0,8.9);
\fill[green!70!black] (17.1,8.95) rectangle (17.35,8.85);
\node[font=\sffamily\tiny, black!60, anchor=west] at (17.45,8.9) {Llama3-8B};

\draw[black!20] (13.6,8.65) -- (19.8,8.65);

\node[font=\sffamily\tiny\bfseries, black!60, anchor=west] at (13.7,8.4) {Aggregate:};

\node[font=\sffamily\tiny, green!70!black, anchor=west] at (13.7,8.0) {Economy (.75)};
\fill[green!70!black] (15.7,8.15) rectangle (18.0,7.95);
\fill[red!50] (18.0,8.15) rectangle (18.8,7.95);
\fill[orange!70] (18.8,8.15) rectangle (19.1,7.95);
\node[font=\sffamily\tiny, black!60, anchor=west] at (15.7,7.75) {acc: 76.8\% | cost: 94\% $\downarrow$};

\node[font=\sffamily\tiny, orange!70, anchor=west] at (13.7,7.4) {Standard (.88)};
\fill[green!70!black] (15.7,7.55) rectangle (17.2,7.35);
\fill[red!50] (17.2,7.55) rectangle (18.5,7.35);
\fill[orange!70] (18.5,7.55) rectangle (19.1,7.35);
\node[font=\sffamily\tiny, black!60, anchor=west] at (15.7,7.15) {acc: 88.5\% | cost: 74\% $\downarrow$};

\node[font=\sffamily\tiny, red!70, anchor=west] at (13.7,6.8) {Premium (.95)};
\fill[green!70!black] (15.7,6.95) rectangle (16.0,6.75);
\fill[red!50] (16.0,6.95) rectangle (17.3,6.75);
\fill[orange!70] (17.3,6.95) rectangle (19.1,6.75);
\node[font=\sffamily\tiny, black!60, anchor=west] at (15.7,6.55) {acc: 95.2\% | cost: 15\% $\downarrow$};

\fill[green!70!black] (13.7,6.3) rectangle (13.9,6.2);
\node[font=\sffamily\tiny, black!50, anchor=west] at (13.95,6.25) {cheap};
\fill[red!50] (14.8,6.3) rectangle (15.0,6.2);
\node[font=\sffamily\tiny, black!50, anchor=west] at (15.05,6.25) {mid};
\fill[orange!70] (15.7,6.3) rectangle (15.9,6.2);
\node[font=\sffamily\tiny, black!50, anchor=west] at (15.95,6.25) {frontier};

\fill[cyan!15, rounded corners=3pt] (13.6,5.7) rectangle (19.9,6.0);
\node[font=\sffamily\scriptsize\bfseries, cyan!70!black, align=center] at (16.75,5.85) {Floor compliance: 100\% \textbar\ Same trained model};

\fill[white, rounded corners=6pt] (20.5,5.7) rectangle (26.5,10.2);
\draw[black!20, thick, rounded corners=6pt] (20.5,5.7) rectangle (26.5,10.2);
\node[font=\sffamily\scriptsize\bfseries, black!60] at (23.5,9.95) {Learned Constraint Mechanism};

\fill[orange!20, rounded corners=3pt] (20.9,9.4) rectangle (22.2,9.7);
\node[font=\sffamily\tiny\bfseries, orange!70!black] at (21.55,9.55) {$\tau$ input};

\fill[red!15, rounded corners=3pt] (22.6,9.4) rectangle (23.9,9.7);
\node[font=\sffamily\tiny\bfseries, red!60!black] at (23.25,9.55) {$\lambda$ dual};

\fill[cyan!20, rounded corners=3pt] (24.3,9.4) rectangle (25.6,9.7);
\node[font=\sffamily\tiny\bfseries, cyan!70!black] at (24.95,9.55) {$\mu$ output};

\draw[-{Stealth[length=1.5mm]}, thick, black!40] (22.2,9.55) -- (22.6,9.55);
\draw[-{Stealth[length=1.5mm]}, thick, black!40] (23.9,9.55) -- (24.3,9.55);

\node[font=\sffamily\tiny, black!50, align=center] at (21.55,9.0) {target\\accuracy};
\node[font=\sffamily\tiny, black!50, align=center] at (23.25,9.0) {constraint\\pressure};
\node[font=\sffamily\tiny, black!50, align=center] at (24.95,9.0) {quality\\preference};

\node[font=\sffamily\tiny\bfseries, black!60] at (22.5,8.2) {Learned Translation};

\draw[-{Stealth[length=1.5mm]}, thick, black!30] (21.2,6.4) -- (21.2,7.9);
\draw[-{Stealth[length=1.5mm]}, thick, black!30] (21.2,6.4) -- (23.8,6.4);
\node[font=\sffamily\tiny, black!40, anchor=east] at (21.1,7.2) {$\mu$};
\node[font=\sffamily\tiny, black!40, anchor=west] at (23.83,6.4) {$\tau$};

\draw[very thick, cyan!60] (21.5,6.6) -- (23.5,7.7);

\fill[green!70!black] (21.7,6.696) circle (0.08);
\fill[yellow!70!orange] (22.5,7.143) circle (0.08);
\fill[red!50] (23.3,7.59) circle (0.08);

\node[font=\sffamily\tiny, cyan!60!black] at (22.8,7.85) {$r$=0.97};

\fill[gray!5, rounded corners=3pt] (23.8,6.7) rectangle (26.4,8.5);
\node[font=\sffamily\tiny, black!70, align=left, anchor=north west, text width=2.4cm] at (23.9,8.4) {%
\textbf{How it works:}\\[2pt]
$\tau \uparrow$ $\rightarrow$ $\lambda$ learns pressure\\[2pt]
$\lambda \uparrow$ $\rightarrow$ policy outputs $\mu \uparrow$\\[2pt]
$\mu \uparrow$ $\rightarrow$ favor quality models
};

\fill[cyan!10, rounded corners=3pt] (20.7,5.85) rectangle (26.3,6.15);
\node[font=\sffamily\tiny\bfseries, cyan!60!black] at (23.5,6.0) {Result: acc $\geq \tau$ guaranteed};

\fill[blue!8, rounded corners=10pt] (-1.0,0.0) rectangle (20.8,4.2);
\node[grouplabel, blue!70!black] at (9.9,3.75) {Inference Pipeline};

\fill[yellow!20, rounded corners=10pt] (21.1,-1.5) rectangle (27.0,4.2);
\node[grouplabel, black] at (24.05,3.85) {Model Pool};

\node[input] (query) at (0,2.2) {Query $x$};
\node[input, fill=orange!25, draw=orange!80] (tau) at (0,1.0) {Target $\tau$};

\node[component, fill=yellow!20] (encoder) at (3.2,2.2) {DeBERTa-v3\\(22M params)};

\node[smallcomp, fill=yellow!10] (embed) at (5.9,2.2) {$z \in \mathbb{R}^{256}$};

\node[component, fill=orange!15] (condition) at (8.6,2.2) {$(\tau, \lambda)$\\Conditioning};
\node[smallcomp, fill=red!15, draw=red!60] (lambda) at (8.6,0.7) {$\lambda$};
\node[complabel] at (9.85,0.7) {\textit{dual var}};

\node[component, fill=cyan!15] (policy) at (11.6,2.2) {Beta Policy\\$\pi_\theta(\mu|x,\tau)$};

\node[circlenode, fill=cyan!20, draw=cyan!70] (mu) at (13.8,2.2) {$\mu$};

\node[component, fill=green!12, minimum width=3.5cm, minimum height=1.1cm] (scoring) at (16.6,2.2) {
\begin{tabular}{c}
\textbf{Routing Score}\\[1pt]
$s_i = p_i + \mu b_i - (1{-}\mu)^\gamma c_i$
\end{tabular}
};

\node[smallcomp, fill=gray!15] (argmax) at (19.8,2.2) {$\arg\max_i$};

\node[modelbox, fill=green!15, minimum width=2.0cm] (m1) at (22.4,3.3) {Llama-1B};
\node[modelbox, fill=green!12, minimum width=2.0cm] (m2) at (25.7,3.3) {Llama-8B};
\node[modelbox, fill=green!10, minimum width=2.0cm] (m3) at (22.4,2.55) {Llama-70B};
\node[modelbox, fill=yellow!18, minimum width=2.0cm] (m4) at (25.7,2.55) {Mistral-7B};
\node[modelbox, fill=yellow!15, minimum width=2.0cm] (m5) at (22.4,1.8) {Mixtral-8x7B};
\node[modelbox, fill=orange!12, minimum width=2.0cm] (m6) at (25.7,1.8) {GPT-3.5};
\node[modelbox, fill=orange!18, minimum width=2.0cm] (m7) at (22.4,1.05) {GPT-4o};
\node[modelbox, fill=red!15, minimum width=2.0cm] (mk) at (25.7,1.05) {Claude-3.5};
\node[modelbox, fill=red!12, minimum width=2.0cm] (m9) at (22.4,0.3) {o3-mini};
\node[modelbox, fill=red!20, minimum width=2.0cm] (m10) at (25.7,0.3) {GPT-4};
\node[font=\sffamily\scriptsize, black!50] at (22.4,-0.35) {$\vdots$};
\node[font=\sffamily\scriptsize, black!50] at (25.7,-0.35) {$\vdots$};
\node[font=\sffamily\tiny, black!60] at (24.05,-0.85) {$K$ models available};

\draw[arrow] (query) -- (encoder);
\draw[arrow] (encoder) -- (embed);
\draw[arrow] (embed) -- (condition);
\draw[arrow] (condition) -- (policy);
\draw[arrow] (policy) -- (mu);
\draw[arrow] (mu) -- (scoring);
\draw[arrow] (scoring) -- (argmax);

\draw[-{Stealth[length=2mm]}, thick, orange!70] (tau.east) -- (7.25,1.0) -- (7.25,1.9) -- (7.5,1.9);

\draw[arrow, red!60] (lambda) -- (condition);

\draw[thick, black!70] (argmax.east) -- (21.0,2.2) -- (21.0,3.3);
\draw[thick, black!70] (21.0,2.2) -- (21.0,0.3);
\draw[arrow] (21.0,3.3) -- (m1.west);
\draw[arrow] (21.0,2.55) -- (m3.west);
\draw[arrow] (21.0,1.8) -- (m5.west);
\draw[arrow] (21.0,1.05) -- (m7.west);
\draw[arrow] (21.0,0.3) -- (m9.west);

\fill[orange!10, rounded corners=10pt] (2.0,-3.7) rectangle (18.5,-0.2);
\node[grouplabel, orange!70!black] at (10.25,-3.4) {Training: Primal-Dual Constraint Enforcement};

\node[smallcomp, fill=gray!15] (batch) at (16.4,-1.8) {$\bar{p}_{\text{batch}}$};
\node[complabel] at (16.4,-2.55) {\textit{batch accuracy}};

\node[component, fill=orange!12, draw=orange!50, minimum width=2.6cm] (check) at (12.2,-1.8) {$\bar{p} \geq \tau$?};
\node[complabel] at (12.2,-2.55) {\textit{constraint check}};

\node[component, fill=orange!18, draw=orange!60, minimum width=4.2cm] (dualupdate) at (6.0,-1.8) {$\lambda \gets [\lambda + \eta_\lambda(\tau - \bar{p})]_+$};
\node[complabel] at (6.0,-2.55) {\textit{dual variable update}};

\draw[arrow, black!40] (24.05,-1.2) -- (24.05,-1.8) -- (batch.east);
\draw[arrow, black!40] (batch) -- (check);
\draw[arrow, black!40] (check) -- (dualupdate);

\draw[-{Stealth[length=2mm]}, thick, orange!50] (tau.south) -- (0,-1.8) -- (dualupdate.west);

\draw[feedbackarrow] (dualupdate.north) -- ++(0,0.6) -| (lambda.south);
\node[complabel, red!60] at (7.1,-0.95) {\textit{feedback}};

\end{tikzpicture}%
}
\caption{\textbf{PROTEUS System Architecture.} \textit{Top:} Runtime adaptation showing multi-tier query streams where each query arrives with $\tau$ derived from customer SLA tier. The same trained model serves all tiers simultaneously. Per-query adaptation routes based on both target accuracy and query difficulty (Premium easy questions use cheap models; Premium hard questions use expensive models). \textit{Middle:} The inference pipeline encodes queries via DeBERTa-v3; the embedding feeds both a performance prediction head and the $\tau$-conditioned policy that outputs quality preference $\mu \in [0,1]$. The routing score combines predicted performance $p_i$, learned boosts $b_i$, and non-linear cost weighting with learnable $\gamma \in [2, 8]$. \textit{Bottom:} During training, session-based learning fixes a single $\tau$ per session (multiple batches) to provide stable constraint signals. Batch accuracy drives Lagrangian dual updates; the feedback loop injects $\lambda$ into the policy to teach the $\tau \rightarrow \mu$ mapping.}
\label{fig:architecture}
\end{figure*}

\subsection{System Architecture}
\textbf{Query Encoding.} Because routing decisions must be fast and the encoder adds latency to every query, we prioritize efficiency by using DeBERTa-v3-small~\citep{debertav3} for query encoding. This 22M-parameter model matches larger models like RoBERTa-base on natural language understanding (NLU) benchmarks while running 5$\times$ faster than RoBERTa-base's 125M parameters. The encoder projects queries to a 256-dimensional embedding $z$ via a 2-layer MLP, and this embedding feeds two heads: the $\tau$-conditioned policy network that outputs $\mu$, and a performance prediction head that outputs $p_i(x)$ for each model. These heads are intentionally independent. The performance head $p_i(x)$ captures \textit{model capability} (how well does model $i$ handle this query?), while $\mu$ captures \textit{operator intent} (how much quality versus cost?). Coupling them would incorrectly make accuracy predictions depend on $\tau$. Instead, $\tau$ influences only the quality-cost preference, with the scoring function combining both signals. On a single A100 GPU, encoding adds less than 2ms per query.

\textbf{Quality Preference Output.} Rather than outputting a probability distribution over models (which would couple selection directly to $\tau$ and require $K$-way exploration), the policy outputs a continuous quality preference $\mu \in [0,1]$. This decouples the quality-cost tradeoff decision from the model selection mechanism. Here $\mu$ represents how much to prioritize quality versus cost, while the scoring function (Equation~\ref{eq:routing}) translates this preference into model selection, enabling smooth interpolation across operating points and efficient 1-dimensional exploration. When $\mu \approx 0$ the router favors cheap models, and when $\mu \approx 1$ it favors expensive, high-quality models, with the policy learning to map $(\text{query}, \tau)$ to appropriate $\mu$ values. We use a Beta distribution for $\mu$ rather than a sigmoid output because Beta provides learnable concentration. During training, lower concentration encourages exploration across the quality-cost spectrum, while at inference the distribution concentrates around the learned preference. A categorical distribution over $K$ models would require $K$-way exploration and lack the smooth interpolation that continuous $\mu$ provides. During training, a dual variable $\lambda$ (described below) provides additional constraint feedback.

\textbf{Adaptive Routing Scores.} Given $\mu$, model selection uses this scoring function.
\begin{equation}
s_i = p_i(x) + \mu \cdot b_i - (1-\mu)^{\gamma} \cdot c_i
\label{eq:routing}
\end{equation}
The formula combines three signals. First, $p_i(x)$ ensures capable models are always considered regardless of $\tau$. Second, the $\mu \cdot b_i$ term boosts quality-advantaged models when the policy seeks quality. Third, the $(1{-}\mu)^\gamma \cdot c_i$ term penalizes expensive models when the policy seeks cost reduction. The router selects $m^* = \arg\max_i s_i$, where $p_i(x)$ is predicted model performance from a shared linear layer that maps the query embedding to $K$ correctness probabilities via sigmoid activation (one per model), $b_i$ is a learned per-model quality boost that captures systematic model strengths independent of query (while $p_i(x)$ varies per-query, $b_i$ encodes that certain models have consistent quality advantages that should be weighted when $\mu$ favors quality), and $\gamma$ controls cost sensitivity. The non-linear term $(1-\mu)^\gamma$ is important because, unlike linear scoring~\citep{carrot2025}, it lets the policy learn dataset-specific cost-quality tradeoffs. We make $\gamma$ learnable within 2--8, where the bounds have geometric interpretation: at $\gamma{=}2$ the cost term $(1{-}\mu)^2$ is quadratic in quality preference, while at $\gamma{=}8$ it becomes nearly binary (costs matter only when $\mu{\approx}0$). This lets the system discover appropriate sensitivity. The ablation (Table~\ref{tab:ablation}) shows fixing $\gamma{=}3.0$ hurts SPROUT ($-$0.62\% accuracy) where cost variation is 9.5$\times$ wider than RouterBench.

\subsection{Constraint Enforcement via Learned Dual Variables}
The core challenge is ensuring the router actually meets accuracy targets, since simply adding $\tau$ as input does not guarantee the output respects it. We use Lagrangian dual variables~\citep{altman1999cmdp,tessler2019reward} to enforce constraints during training by injecting constraint feedback directly into the policy.

\textbf{The Feedback Loop.} During training, we track batch accuracy $\bar{p}_{\text{batch}}$ and compare it against the target $\tau$, with a dual variable $\lambda$ adjusting based on constraint violations.
\begin{equation}
\lambda_{t+1} = \left[\lambda_t + \eta_\lambda \cdot (\tau - \bar{p}_{\text{batch}})\right]_+
\label{eq:dual_update}
\end{equation}
When accuracy falls below target ($\bar{p} < \tau$), $\lambda$ increases to penalize the policy for cost-seeking behavior, and when accuracy exceeds target, $\lambda$ decreases to allow more aggressive cost optimization. The $[\cdot]_+$ operator ensures $\lambda \geq 0$, and during training this creates a feedback loop where the policy learns to associate $\tau$ values with appropriate routing behavior.
Crucially, $\lambda$ is injected into the policy network during training, so the policy sees $\lambda$ as input alongside the query and $\tau$ and learns to anticipate constraint pressure. When $\lambda$ is high the policy outputs higher $\mu$ values that favor quality, and when $\lambda$ is low it outputs lower $\mu$ values that favor cost, creating correlation between $\tau$ and $\mu$. Because high targets produce high $\lambda$ values during training, the policy learns to output high $\mu$ for high $\tau$. At inference time, $\lambda$ is fixed at 1.0 and the policy responds to $\tau$ alone, since the training phase has already taught it the $\tau \rightarrow \mu$ mapping through $\lambda$-mediated feedback.

\textbf{Policy Training.} We train using Proximal Policy Optimization (PPO)~\citep{ppo2017}, a stable policy gradient method that clips updates to prevent large destabilizing changes. Supervised learning is unsuitable here because routing lacks ground-truth labels. We observe per-model outcomes (model $i$ scores $p_i$ on query $x$) but not which model the router \textit{should} pick, and the optimal choice depends on the cost-accuracy tradeoff specified by $\tau$, which varies at runtime. RL handles this naturally, as the policy explores routing decisions, receives reward feedback, and learns to maximize the objective without explicit supervision. The reward follows the standard Lagrangian formulation for constrained MDPs~\citep{altman1999cmdp,stooke2020responsive}:
\begin{equation}
r(x, \mu) = w_q(\tau) \cdot p_{m^*}(x) - w_c(\tau) \cdot \hat{c}_{m^*} + \lambda(p_{m^*}(x) - \tau)
\label{eq:reward}
\end{equation}
where $w_q(\tau) = e^{2\tau_{\text{norm}}}$ and $w_c(\tau) = e^{2(1-\tau_{\text{norm}})}$ are exponential weights that shift emphasis from cost minimization at low $\tau$ to quality maximization at high $\tau$, with $\tau_{\text{norm}} = (\tau - \tau_{\min})/(\tau_{\max} - \tau_{\min})$. Because accuracy lies in $[0,1]$ while raw costs span orders of magnitude across model pools, we normalize costs online using exponential moving average of percentile-based bounds, yielding $\hat{c} \in [0,1]$ to ensure comparable reward magnitudes. This reward function provides the training signal for RL and differs from the routing score $s_i$ (Equation~\ref{eq:routing}) used at inference.
We employ session-based training where each session fixes a single $\tau$ value sampled uniformly from the target range and trains for multiple batches before sampling a new $\tau$. This staged approach helps the policy learn stable $\tau$-conditioned behavior by providing consistent constraint signals within each session while ensuring uniform coverage across the full $\tau$ spectrum over the course of training.

\textbf{Training Configuration.} We use batch size 32 to balance gradient stability with memory constraints on a single A100 GPU, and training runs for 10K steps (roughly 4 hours per dataset), which is sufficient for dual variable convergence as monitored on validation accuracy. The dual learning rate $\eta_\lambda{=}0.4$ follows the ``faster dual'' heuristic~\citep{stooke2020responsive}, which recommends the multiplier adapt orders of magnitude faster than the policy (here ${\sim}10^3{\times}$ faster than the policy rate of $3{\times}10^{-4}$). We determined 0.4 empirically. Values above 1.0 caused $\lambda$ to spike erratically (0$\rightarrow$1.8 within hundreds of steps), destabilizing training, while values below 0.1 slowed constraint convergence.

\section{Experiments}
\label{sec:experiments}

\subsection{Setup}
\textbf{Datasets.} Evaluating LLM routers requires benchmarks that provide per-query correctness labels across multiple models with realistic cost signals, which standard NLP benchmarks lack. We evaluate on the two gold-standard routing benchmarks that meet these requirements.
\textbf{RouterBench}~\citep{routerbench2024} provides 405K inference outcomes across 11 models, spanning open-source variants like Llama-2-7B/13B/70B, Mistral-7B, and Mixtral-8x7B as well as proprietary models including GPT-3.5, GPT-4, and Claude-2. Tasks cover reasoning with MMLU and HellaSwag, mathematics with GSM8K and MATH, and coding with HumanEval and MBPP, with model costs ranging from \$0.0001 to \$0.01 per query.
\textbf{SPROUT}~\citep{carrot2025} complements RouterBench with 45K queries across 14 models, including frontier systems like GPT-4o, Claude-3.5-Sonnet, and o3-mini, with Llama variants ranging from 1B to 405B parameters. It uses instruction-following queries with LLM-based evaluation, and costs span \$0.0001 to \$0.05 per query.
Together, these benchmarks test complementary aspects: RouterBench provides scale and task diversity with established models, while SPROUT tests generalization to modern model pools with extreme cost variation. Oracle accuracy is 91.4\% on RouterBench and 98.6\% on SPROUT. We evaluate $\tau \in [0.85, 0.91]$ for RouterBench and $\tau \in [0.85, 0.95]$ for SPROUT, using 70/15/15 train/val/test splits for RouterBench and official splits for SPROUT.

\textbf{Baselines.} We compare against static strategies, learned routers, and ablated variants. For static baselines, Always-Best routes all queries to the highest-performing model (GPT-4 for RouterBench, o3-mini for SPROUT), maximizing accuracy but ignoring cost. Always-Cheapest routes to the lowest-cost model (Llama-2-7B for RouterBench, Llama-1B for SPROUT). Random provides uniform model selection. For learned routers, KNN~\citep{routerbench2024} uses 5-nearest-neighbors to predict mean model performance via collaborative filtering from query embeddings. MLP~\citep{routerbench2024} uses 3-layer networks for the same task. CARROT~\citep{carrot2025} uses RoBERTa-base (125M parameters) to predict best model with score$_i = (1{-}\mu) \cdot \text{acc}_i - \mu \cdot \text{cost}_i$ where we evaluate $\mu \in \{0.1, 0.5, 0.9\}$ representing quality-focused, balanced, and cost-focused configurations. RouteLLM-Style~\citep{routellm2024} extends binary win-rate prediction to multi-model selection by training a classifier predicting P(model$_i$ is best $|$ query). OmniRouter-Style~\citep{omnirouter2025} adapts batch-level Lagrangian optimization to per-query routing with dual-head accuracy prediction and constraint parameter $\alpha$. For ablations, we test PROTEUS without $\tau$-conditioning (Unconstrained), fixed $\gamma$=3.0 (Fixed-Gamma), and no critic baseline (NoCritic) to isolate each component's contribution.

\textbf{Metrics.} We evaluate using standard routing metrics~\citep{routerbench2024} (accuracy, cost, oracle gap) alongside metrics that capture target-driven capabilities.
\textit{Adaptability Metrics.} \textbf{$\tau$-$\mu$ Correlation} measures Pearson correlation between requested target $\tau$ and policy output $\mu$. High correlation ($>$0.9) indicates the policy faithfully translates requirements into routing behavior. \textbf{SLA Compliance} measures the percentage of $\tau$ levels where achieved accuracy meets or exceeds $\tau$. We also report tolerance-band compliance at $\pm$2\% and $\pm$5\% thresholds.
\textit{Efficiency Metrics.} We introduce Routing Efficiency (RE) and Routing Performance Index (RPI) to capture routing efficiency holistically. \textbf{RE} measures accuracy gain per unit latency: $\text{RE} = \frac{\text{Acc} - \text{Acc}_{\text{random}}}{\text{Latency}}$ (pp/ms). Higher RE indicates the router extracts more accuracy improvement per millisecond of computational overhead. A router with RE of 10 pp/ms gains 10 percentage points of accuracy over random selection for each millisecond spent on routing decisions. \textbf{RPI} balances quality, cost, and latency: $\text{RPI} = \frac{\text{Acc}}{\text{Acc}_{\text{oracle}}} \times \left(1 - \frac{\text{Cost}}{\text{Cost}_{\text{max}}}\right) \times \left(1 - \frac{t_{\text{router}}}{t_{\text{LLM}}}\right) \times 100$. RPI combines three normalized factors into a single score where 100 represents ideal routing (oracle accuracy, zero cost, zero latency). This composite metric penalizes routers that achieve high accuracy through expensive models or slow inference, rewarding efficient cost-quality tradeoffs.

\textbf{Implementation.} We implement PROTEUS in PyTorch with HuggingFace Transformers for DeBERTa-v3-small encoding. The policy uses a 2-layer MLP (256 hidden units) mapping query embeddings, $\tau$, and $\lambda$ to Beta parameters via softplus, with $\lambda$-gating for constraint-aware attention. Training uses AdamW (lr=$3{\times}10^{-4}$, batch=32, 10K steps), dual updates every 5 batches ($\eta_\lambda$=0.4), gradient clipping (max norm 1.0), completing in 4 hours per dataset on one A100 GPU. The router integrates with vLLM~\citep{vllm2023} and load balancers~\citep{llmloadbalancer2025} without modification.

\section{Results}
\label{sec:results}

\subsection{Runtime Adaptability: The Core Capability}

\begin{table}[t]
\centering
\caption{\textbf{PROTEUS Runtime Adaptability.} A single trained policy accepts accuracy targets $\tau$ at runtime and adapts routing accordingly. Baselines require either retraining per-target (impractical for dynamic SLAs) or parameter sweeping (which yields arbitrary accuracy, not the requested target).}
\label{tab:proteus_adaptive}
\small
\begin{tabular}{lcc}
\toprule
\textbf{Adaptability Metric} & \textbf{RouterBench} & \textbf{SPROUT} \\
\midrule
$\tau$-$\mu$ Correlation & 0.973 & 0.981 \\
Floor Compliance (acc $\geq \tau$) & 100\% & 100\% \\
SLA Compliance ($\pm$5\% tol.) & 100\% & 67\%$^\dagger$ \\
SLA Compliance ($\pm$2\% tol.) & 44\% & 11\%$^\dagger$ \\
Cost Range (min/max $\tau$) & 3.7$\times$ & 9.5$\times$ \\
\bottomrule
\end{tabular}
\vspace{0.3em}
\hrule
\vspace{0.2em}
\textit{Performance by Service Tier}
\vspace{0.1em}

\small
\begin{tabular}{llccc}
\toprule
\textbf{Dataset} & \textbf{Tier} & \textbf{$\tau$ Range} & \textbf{Accuracy (margin)} & \textbf{Cost} \\
\midrule
RouterBench & Economy & 0.85--0.87 & 88.8\% (+2.8\%) & \$0.20 \\
& Standard & 0.87--0.89 & 90.4\% (+2.4\%) & \$0.29 \\
& Premium & 0.89--0.91 & 91.3\% (+1.3\%) & \$0.50 \\
\midrule
SPROUT & Economy & 0.85--0.88 & 92.3\% (+6.3\%) & \$0.30 \\
& Standard & 0.88--0.92 & 93.8\% (+5.8\%) & \$0.62 \\
& Premium & 0.92--0.95 & 96.0\% (+4.0\%) & \$1.69 \\
\bottomrule
\end{tabular}
\vspace{0.2em}

{\small \textit{Note:} Costs in \$/1K queries. Margin = achieved $-$ $\tau$ floor. $^\dagger$Lower tolerance-band compliance reflects intentional overshooting (+6\% margin at low $\tau$ exceeds $\pm$5\% band).}
\end{table}

\begin{figure*}[t]
\centering
\includegraphics[width=\textwidth]{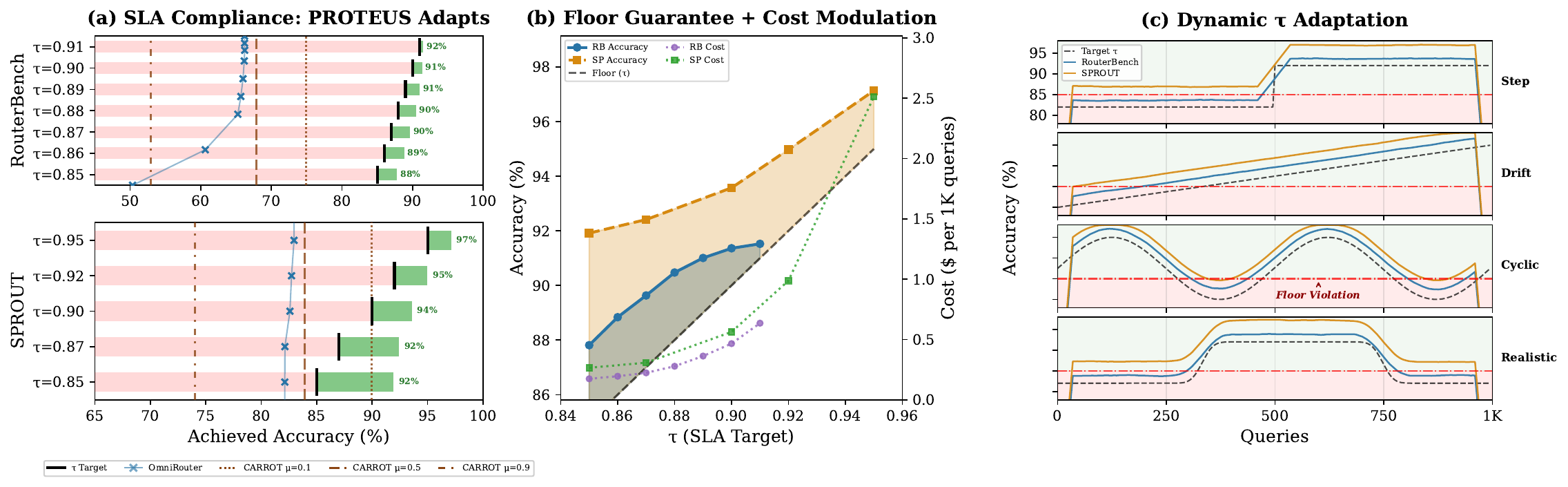}
\caption{\textbf{Main Results.} \textbf{(a) SLA Compliance:} PROTEUS (bars) consistently meets or exceeds each $\tau$ target (black lines), while baselines fail. OmniRouter (blue X markers) plateaus below targets despite per-$\tau$ training; CARROT (brown dotted lines) achieves fixed accuracy regardless of $\tau$. Green portions show accuracy above target. \textbf{(b) Floor Guarantee + Cost:} Achieved accuracy for RouterBench (blue) and SPROUT (orange) exceeds the floor constraint (dashed diagonal) across all $\tau$ values. Cost (dotted lines, right axis) increases with $\tau$. Shaded regions show the reliability margin. \textbf{(c) Dynamic Adaptation:} Four $\tau$-change scenarios showing PROTEUS tracking varying targets in real-time. Both datasets follow the target $\tau$ (dashed) with minimal lag across step, drift, cyclic, and realistic patterns.}
\label{fig:main_results}
\end{figure*}

Table~\ref{tab:proteus_adaptive} quantifies PROTEUS's runtime adaptability, which is the ability to accept accuracy targets $\tau$ at inference time and adapt routing accordingly. The upper portion reports adaptability metrics across all $\tau$ values, while the lower portion shows concrete service tier configurations where baselines fail when evaluated on SLA compliance.

\textbf{Baselines fail to meet accuracy targets.} Figure~\ref{fig:main_results}(a) compares PROTEUS against OmniRouter~\citep{omnirouter2025} and CARROT~\citep{carrot2025} on SLA compliance. OmniRouter (blue X markers) achieves only 22\% floor compliance on RouterBench and 0\% on SPROUT despite being trained with Lagrangian constraints, and it cannot adapt to runtime $\tau$ changes. CARROT (brown dotted lines) produces fixed accuracy regardless of the requested target, with $\mu$ values of 0.1, 0.5, and 0.9 yielding 74.9\%, 67.9\%, and 52.9\% accuracy on RouterBench respectively, none of which match the requested $\tau$.

\textbf{Why not retrain or sweep parameters?} Two alternatives exist: retraining a separate model for each target $\tau$, or sweeping baseline parameters like $\alpha$ or $\mu$ to approximate desired accuracy. Neither works well. Retraining per-$\tau$ requires maintaining $N$ models for $N$ operating points and cannot adapt at runtime, so if a customer's SLA changes mid-session or system load spikes, there is no mechanism to respond. Parameter sweeping is equally problematic because CARROT's~\citep{carrot2025} $\mu$ controls the cost-quality tradeoff rather than the accuracy target. Setting $\mu$=0.5 yields 67.9\% accuracy on RouterBench, but if an operator needs exactly 85\%, no $\mu$ value guarantees it. PROTEUS solves this by conditioning on $\tau$ directly, allowing one model to serve arbitrary targets with runtime adaptation.

\textbf{PROTEUS learns target-aware routing.} The $\tau$-$\mu$ correlation reaches 0.973 on RouterBench and 0.981 on SPROUT (Table~\ref{tab:proteus_adaptive}), demonstrating that PROTEUS faithfully translates accuracy targets into quality preferences. A correlation near 1.0 means that when operators request higher $\tau$, the policy reliably outputs higher $\mu$ values that favor expensive, high-quality models, and this learned translation is what allows runtime adaptability.

\textbf{Floor guarantee with reliability margin.} On held-out test data, PROTEUS achieves floor compliance across all evaluated $\tau$ levels, meaning achieved accuracy meets or exceeds $\tau$. Figure~\ref{fig:main_results}(b) shows PROTEUS consistently exceeds the floor with a reliability margin. For the Economy tier where $\tau$ ranges from 0.85 to 0.87, PROTEUS achieves 88.8\% on RouterBench (2.8\% above the tier floor) and 92.3\% on SPROUT (6.3\% above the floor). This overshoot is intentional because the learned Lagrangian multiplier $\lambda$ provides a buffer against distribution shift. The margin shrinks as $\tau$ approaches the oracle ceiling of 91.4\% on RouterBench and 98.6\% on SPROUT, showing that the policy recognizes feasibility limits.

\textbf{Tiered service delivery.} The lower portion of Table~\ref{tab:proteus_adaptive} shows practical deployment configurations where each dataset has different $\tau$ ranges reflecting its oracle ceiling. RouterBench tops out at 91.4\%, so Premium uses $\tau \in [0.89, 0.91]$, while SPROUT reaches 98.6\%, enabling Premium at $\tau \in [0.92, 0.95]$. On RouterBench, Economy tier costs \$0.20/1K queries and delivers 88.8\% accuracy, whereas Premium costs \$0.50/1K and achieves 91.3\%, representing a 2.5$\times$ cost increase for 2.5 percentage points of accuracy. On SPROUT, the cost-accuracy gradient is steeper, with Economy at \$0.30/1K and Premium at \$1.69/1K, a 5.6$\times$ increase for a 3.7 point accuracy gain. This cost-accuracy tradeoff enables operators to offer differentiated SLAs.

\textbf{Dynamic adaptation.} Figure~\ref{fig:main_results}(c) evaluates PROTEUS when $\tau$ changes mid-session across four scenarios: step change (sudden SLA upgrades from $\tau$=0.82 to 0.92), gradual drift (linear increase from 0.80 to 0.95), cyclic (sinusoidal variation modeling time-of-day pricing), and realistic (off-peak and peak hour patterns with smoothed transitions). Each scenario spans 1,000 queries with 5 random seeds, and PROTEUS tracks all patterns. Floor satisfaction reaches 77\% on RouterBench and ranges from 82\% to 86\% on SPROUT, with RouterBench showing 1.6 to 2.3\% overshoot and SPROUT showing 4.3 to 5.2\% overshoot. Critically, there is zero adaptation delay because $\tau$ is a direct input, so the policy responds instantaneously without retraining or parameter tuning. Operators can adjust targets per-query, per-customer, or based on system load.

\subsection{Standard Routing Performance}

Beyond adaptability, we evaluate PROTEUS on standard routing metrics. Table~\ref{tab:main_results} presents the comparison across both datasets.

\begin{table}[htbp]
\centering
\caption{\textbf{Routing Performance Comparison.} PROTEUS achieves near-oracle accuracy with significant cost reduction and highest routing efficiency. \textit{RB}=RouterBench, \textit{SP}=SPROUT.}
\label{tab:main_results}
\small
\renewcommand{\arraystretch}{0.85}
\setlength{\tabcolsep}{3pt}
\begin{tabular}{l cc cc cc cc}
\toprule
& \multicolumn{2}{c}{\textbf{Acc. (\%)}} & \multicolumn{2}{c}{\textbf{Cost (\$/1K)}} & \multicolumn{2}{c}{\textbf{Routing Eff.}} & \multicolumn{2}{c}{\textbf{Overall Perf.}} \\
& \multicolumn{2}{c}{} & \multicolumn{2}{c}{} & \multicolumn{2}{c}{\scriptsize (RE $\uparrow$)} & \multicolumn{2}{c}{\scriptsize (RPI $\uparrow$)} \\
\cmidrule(lr){2-3} \cmidrule(lr){4-5} \cmidrule(lr){6-7} \cmidrule(lr){8-9}
Method & RB & SP & RB & SP & RB & SP & RB & SP \\
\midrule
\multicolumn{9}{l}{\textit{Static Baselines}} \\
Random & 52.4 & 71.8 & 0.86 & 1.7 & N/A$^\ddagger$ & N/A$^\ddagger$ & 42.4 & 56.8 \\
Cheapest & 30.4 & 63.5 & 0.05 & 0.06 & N/A$^\ddagger$ & N/A$^\ddagger$ & 32.7 & 63.8 \\
Best Fixed$^\dagger$ & 77.9 & 90.7 & 3.3 & 5.5 & N/A$^\ddagger$ & N/A$^\ddagger$ & 0.0 & 25.8 \\
\midrule
\multicolumn{9}{l}{\textit{Learned Routers}} \\
KNN & 77.2 & 73.8 & 2.3 & 1.7 & 7.4 & 0.9 & 25.5 & 58.1 \\
MLP & 76.7 & 77.1 & 2.3 & 1.7 & 7.3 & 2.3 & 25.4 & 60.7 \\
CARROT-Quality & 74.9 & 89.9 & 1.34 & 2.24 & 6.9 & 8.0 & 41.4 & 60.9 \\
CARROT-Balanced & 67.9 & 83.9 & 0.21 & 0.35 & 6.0 & 6.8 & 69.5 & 81.0 \\
CARROT-Cost & 52.9 & 74.0 & \textbf{0.09} & \textbf{0.11} & 0.2 & 1.2 & 55.7 & 69.9 \\
OmniRouter & 66.2 & 82.6 & 0.24 & 0.64 & 5.3 & 6.0 & 67.1 & 76.6 \\
\midrule
\multicolumn{9}{l}{\textit{Ours}} \\
\textbf{PROTEUS} & \textbf{90.1} & \textbf{94.0} & 0.33 & 0.93 & \textbf{11.1} & \textbf{9.5} & \textbf{88.5} & \textbf{83.5} \\
\midrule\midrule
\rowcolor{blue!8} Oracle & 91.4 & 98.6 & 0.39 & 0.60 & N/A$^\ddagger$ & N/A$^\ddagger$ & 88.2 & 92.2 \\
\bottomrule
\end{tabular}
\\[3pt]
\centering
{\scriptsize $^\dagger$Best Fixed = GPT-4 (RB), o3-mini (SP). \quad $^\ddagger$RE undefined for methods with zero routing latency; RPI uses $t_{\text{router}}/t_{\text{LLM}} = 0$. RE computed with batch size 8 latency (Table~\ref{tab:latency}); all learned routers use similar-sized encoders (22--125M params) with comparable latency.}
\end{table}

\textbf{Near-oracle accuracy with cost savings.} PROTEUS achieves 90.1\% accuracy on RouterBench, just 1.3 percentage points below oracle, with cost savings of 90\% versus GPT-4. On SPROUT, PROTEUS achieves 94.0\% accuracy with a 4.6 percentage point gap and 83\% savings versus o3-mini. Learned baselines like KNN and MLP achieve only 77\% accuracy at 7$\times$ higher cost because they default to expensive models rather than learning query-specific routing.

\textbf{Highest routing efficiency.} PROTEUS achieves RE of 11.1 pp/ms on RouterBench and 9.5 pp/ms on SPROUT, meaning each millisecond of routing computation yields more than 10 percentage points of accuracy gain over random selection. On RouterBench, PROTEUS achieves higher RPI than Oracle (88.5 versus 88.2), showing that learned cost-quality tradeoffs can outperform pure quality optimization because accepting a 1.4\% quality reduction saves 15\% cost, yielding a net efficiency gain. On SPROUT, RPI of 83.5 approaches the oracle ceiling of 92.2.

The efficiency advantage stems from two factors. First, PROTEUS uses DeBERTa-v3-small (22M parameters) rather than larger encoders, keeping inference fast while maintaining sufficient representational capacity for routing decisions. Second, the continuous $\mu$ output requires only a single forward pass through the policy network, unlike cascade methods that may invoke multiple models sequentially. Compared to learned baselines using similar-sized encoders (KNN, MLP, CARROT with 22--125M parameters), PROTEUS achieves 50\% higher RE because its $\tau$-conditioning enables query-adaptive routing that matches model capability to query difficulty, whereas fixed-threshold baselines apply uniform routing logic regardless of query characteristics.

\begin{table}[t]
\centering
\caption{\textbf{Router Latency.} Per-query latency (ms) and throughput on A100 GPU. Latency is dominated by the DeBERTa-v3-small encoder; the policy head adds $<$0.1ms.}
\label{tab:latency}
\small
\begin{tabular}{rcccc}
\toprule
\textbf{Batch} & \textbf{p50} & \textbf{p95} & \textbf{p99} & \textbf{Throughput} \\
\midrule
1 & 8.7 & 8.9 & 9.0 & 115 q/s \\
8 & 2.9 & 3.0 & 3.1 & 345 q/s \\
32 & 2.6 & 2.7 & 2.7 & 385 q/s \\
64 & 2.6 & 2.6 & 2.7 & 385 q/s \\
128 & 2.6 & 2.6 & 2.6 & 385 q/s \\
\bottomrule
\end{tabular}
\\[2pt]
\centering
{\scriptsize Throughput = 1000/p50. At batch 8+, $<$3ms latency with 300+ q/s.}
\end{table}

\textbf{Router latency analysis.} Table~\ref{tab:latency} presents microbenchmark results for PROTEUS routing latency across batch sizes. Single-query latency of 8.7ms is negligible compared to LLM inference times of 2--7 seconds, adding less than 0.5\% overhead to end-to-end response time. With batching, per-query latency drops to 2.6ms at batch size 32+, achieving 385 queries/second throughput. The latency is dominated by the DeBERTa-v3-small encoder forward pass, with the policy MLP and scoring function contributing less than 0.1ms combined. This efficiency enables PROTEUS to operate as a lightweight preprocessing step that does not bottleneck the serving pipeline, even under high query loads where batched inference is standard practice.

\subsection{Ablation Study}

\begin{table}[H]
\centering
\caption{\textbf{Ablation Study.} Performance change when removing components from PROTEUS-Full. \textit{$\Delta$Acc}: accuracy change in percentage points. \textit{$\Delta$Cost}: relative cost change (negative values indicate variants are cheaper but achieve lower accuracy). Colors: \colorbox{red!15}{significant} ($|$$\Delta$Acc$|$$>$0.2), \colorbox{green!10}{negligible} ($|$$\Delta$Acc$|$$<$0.1).}
\label{tab:ablation}
\small
\setlength{\tabcolsep}{6pt}
\renewcommand{\arraystretch}{0.9}
\begin{tabular}{l cc cc}
\toprule
& \multicolumn{2}{c}{\textbf{$\Delta$Acc (\%)}} & \multicolumn{2}{c}{\textbf{$\Delta$Cost}} \\
\cmidrule(lr){2-3} \cmidrule(lr){4-5}
Variant & RB & SP & RB & SP \\
\midrule
\textbf{PROTEUS-Full} & \textbf{90.1} & \textbf{94.0} & \$0.33 & \$0.93 \\
\midrule
$-$Constraint ($\lambda$) & \cellcolor{red!15}$-$0.27 & \cellcolor{red!15}$-$0.53 & $-$8\% & $-$25\% \\
$-$Learnable $\gamma$ & \cellcolor{green!10}$\approx$0 & \cellcolor{red!15}$-$0.62 & $\approx$0 & $-$44\% \\
$-$Critic & \cellcolor{green!10}$-$0.04 & \cellcolor{green!10}$-$0.03 & $-$2\% & $-$1\% \\
\bottomrule
\end{tabular}
\vspace{0.1em}

{\small \textit{Takeaway:} $\lambda$ is critical (drops accuracy on both datasets). Learnable $\gamma$ matters only on SPROUT (wider cost range). Critic is optional.}
\end{table}

Table~\ref{tab:ablation} isolates how each architectural component contributes to PROTEUS performance by ablating three elements: the Lagrangian constraint mechanism $\lambda$, the learnable cost sensitivity parameter $\gamma$, and the critic network. Negative cost changes indicate variants route to cheaper models at the expense of accuracy.

\textbf{Constraint mechanism is essential.} Removing $\lambda$ causes the largest accuracy drops of 0.27\% on RouterBench and 0.53\% on SPROUT because without the dual variable feedback loop, the policy learns a fixed quality-cost tradeoff. The 8\% to 25\% cost reduction reveals unconstrained variants sacrifice accuracy to route to cheaper models, validating that learned dual variables enable reliable target satisfaction.
\textbf{Learnable $\gamma$ matters for heterogeneous pools.} Cost sensitivity $\gamma$ in Equation~\ref{eq:routing} has negligible effect on RouterBench but causes a 0.62\% accuracy drop on SPROUT because RouterBench spans only 3.7$\times$ cost variation while SPROUT spans 9.5$\times$ variation. With fixed $\gamma$=3.0, the policy cannot adapt cost weighting to extreme ratios, and the 44\% cost reduction on SPROUT shows the variant defaults to cheaper models.
\textbf{Critic provides marginal benefit.} Removing the critic network yields negligible accuracy changes ($<$0.05\%). This aligns with prior work showing critics provide little benefit for single-step Markov Decision Processes (MDPs)~\citep{schulman2015high}. Since routing is a single-step decision, the REINFORCE gradient estimator suffices.

\FloatBarrier

\section{Conclusion}

PROTEUS represents the first approach to LLM routing that accepts accuracy targets as direct runtime input. Existing routers force operators to tune parameters offline and guess what accuracy might result, but PROTEUS eliminates this indirection by treating accuracy targets as first-class inputs rather than indirect configuration. By conditioning a learned policy on $\tau$ and enforcing constraints via Lagrangian dual variables, a single trained model serves the entire accuracy spectrum from 0.85 to 0.95 without retraining. The $\tau$-$\mu$ correlation reaches 0.97 to 0.98, and floor compliance hits 100\% across all evaluated targets while the closest baseline achieves only 22\%. This performance enables differentiated service delivery that was previously impractical, where premium customers get guaranteed high accuracy while economy tiers optimize for cost, and operators can adjust targets per-query, per-customer, or based on system load with zero adaptation delay.

\textbf{Limitations.} RouterBench's model pool includes Llama-2, GPT-3.5, GPT-4, and Claude-2, all of which predate current frontier models, while SPROUT covers narrower task distributions. The larger oracle gap on SPROUT reflects its higher oracle ceiling at 98.6\% versus RouterBench's 91.4\% combined with wider cost variation at 9.5$\times$ compared to 3.7$\times$, creating more challenging cost-quality tradeoffs. Both datasets exhibit compressed accuracy ranges among mid-tier models, which makes fine-grained $\tau$ distinctions difficult. PROTEUS trains on ground-truth correctness labels, which is the standard setting for routing research~\citep{routerbench2024,carrot2025,routellm2024}, but production systems lack such labels by default, so deployment would require periodic sampling with human annotation or LLM-as-judge~\citep{llm_blender} evaluation to generate training signals. The reliability margin of 1.3 to 6.3\% overshoot buffers against distribution shift, though sustained drift would require policy updates. Additionally, model costs $c_i$ are fixed at training time. If API providers change pricing post-deployment (as frequently occurs with LLM APIs), the learned cost-quality tradeoffs become stale. Short-term mitigation involves re-normalizing costs at inference if relative price rankings are preserved, but significant pricing changes would require retraining, which completes in 4 hours on a single GPU.

\textbf{Future Work.} Several directions warrant investigation. Bandit feedback formulations where only the selected model's outcome is observed would reduce annotation cost and enable continual learning~\citep{llmbandit2025}. Latency-aware routing could jointly optimize accuracy, cost, and response time, relevant for interactive applications where tail latency matters. Multi-objective $\tau$ specifications would allow operators to set accuracy floors and cost ceilings simultaneously, aligning with SLOs that specify both quality and budget constraints. Cost-adaptive routing could treat model costs as runtime inputs rather than static training-time parameters, enabling immediate adaptation to API pricing changes by conditioning the policy on a cost vector $\mathbf{c}$ alongside $\tau$. Reasoning-aware routing extends optimization to jointly select models and allocate reasoning budgets. With reasoning models like o1 and DeepSeek-R1, thinking tokens directly impact both cost and quality, creating a continuous dimension alongside discrete model selection. The policy could output both $\mu$ (model preference) and a reasoning budget $\rho$, enabling finer cost-quality control where the same accuracy target might be achieved via an expensive model with minimal reasoning or a cheaper model with extended thinking. Finally, extending PROTEUS to multi-turn conversations and developing routing benchmarks with diverse model pools spanning wider accuracy ranges remain open challenges.

\section*{Acknowledgments}
We are grateful to the open-source communities behind PyTorch, HuggingFace Transformers, and the DeBERTa model, which made this research possible. We also thank the authors of RouterBench and SPROUT for releasing their datasets and evaluation frameworks publicly, enabling reproducible research in LLM routing. The availability of these resources and the broader culture of open benchmarking in the ML systems community has been invaluable to this work.
Generative AI tools were used to assist with editing and refining the manuscript text. All technical content, experimental design, implementation, and analysis were conducted by the authors, who take full responsibility for the work.

\bibliographystyle{plainnat}
\bibliography{arxiv_ref}

\end{document}